\documentclass[11pt,letterpaper]{article}
\usepackage{naaclhlt2016}
\usepackage{times}
\usepackage{latexsym}
\usepackage{url}
\usepackage{latexsym}

\usepackage{times}
\usepackage{amsmath}
\usepackage{amsthm}
\usepackage{amssymb}
\usepackage{graphicx}
\usepackage{xspace}
\usepackage{tabularx}
\usepackage{multicol}
\usepackage{multirow}
\usepackage{url}
\usepackage{wrapfig}
\usepackage{comment}
\usepackage{listings}
\usepackage[utf8]{inputenc}
\usepackage{fancyvrb}
\usepackage{booktabs}
\usepackage{color,soul}
\usepackage[normalem]{ulem}

\newcommand{\f}{\text{f}}

\newcommand{\bmx}[0]{\begin{bmatrix}}
\newcommand{\emx}[0]{\end{bmatrix}}

\newcommand{\vect}[1]{\mathbf{#1}}
\newcommand{\vects}[1]{\boldsymbol{#1}}
\newcommand{\matr}[1]{\mathbf{#1}}

\newcommand{\vb}[0]{\vect{b}}
\newcommand{\vc}[0]{\vect{c}}

\newcommand{\vh}[0]{\vect{h}}

\newcommand{\vz}[0]{\vect{z}}

\newcommand{\mW}[0]{\matr{W}}

\newcommand{\mE}[0]{\matr{E}}

\newcommand{\TT}[0]{\vects{\theta}}

\newcommand{\LL}[0]{\mathcal{L}}

\newcommand{\RR}[0]{\mathbb{R}}

\newcommand{\ola}{\overleftarrow}
\newcommand{\ora}{\overrightarrow}

\naaclfinalcopy 
\def\naaclpaperid{***} 

\title{Multi-Way, Multilingual Neural Machine Translation \\ with a Shared
Attention Mechanism}

\author{Orhan Firat \\
    Middle East Technical University \\
    {\tt orhan.firat@ceng.metu.edu.tr}
	  \And
	Kyunghyun Cho \\
    New York University \\
	  \And
	Yoshua Bengio \\
    University of Montreal \\
    CIFAR Senior Fellow
}

\date{}

\begin{document}

\maketitle

\begin{abstract}
    We propose multi-way, multilingual neural machine translation. The proposed
    approach enables a single neural translation model to translate between
    multiple languages, with a number of parameters that grows only linearly
    with the number of languages. This is made possible by having a single
    attention mechanism that is shared across all language pairs. We train the
    proposed multi-way, multilingual model on ten language pairs from WMT’15
    simultaneously and observe clear performance improvements over models
    trained on only one language pair. In particular, we observe that the
    proposed model significantly improves the translation quality of
    low-resource language pairs.

\end{abstract}

\section{Introduction}

\paragraph{Neural Machine Translation}

It has been shown that a deep (recurrent) neural network can successfully learn
a complex mapping between variable-length input and output sequences on its own.
Some of the earlier successes in this task have, for instance, been handwriting
recognition~\cite{bottou1997global,graves2009novel} and speech
recognition~\cite{graves2006connectionist,chorowski2015attention}. More
recently, a general framework of encoder-decoder networks has been found to be
effective at learning this kind of sequence-to-sequence mapping by using two
recurrent neural networks \cite{cho2014learning,sutskever2014sequence}.

A basic encoder-decoder network consists of two recurrent networks. The first
network, called an encoder, maps an input sequence of variable length into a
point in a continuous vector space, resulting in a fixed-dimensional context
vector. The other recurrent neural network, called a decoder, then generates a
target sequence again of variable length starting from the context vector. This
approach however has been found to be inefficient in \cite{Cho2014a} when
handling long sentences, due to the difficulty in learning a complex mapping
between an arbitrary long sentence and a single fixed-dimensional vector. 

In \cite{bahdanau2014neural}, a remedy to this issue was proposed by
incorporating an {\em attention mechanism} to the basic encoder-decoder network.
The attention mechanism in the encoder-decoder network frees the network from
having to map a sequence of arbitrary length to a single, fixed-dimensional
vector. Since this attention mechanism was introduced to the encoder-decoder
network for machine translation, neural machine translation, which is purely
based on neural networks to perform full end-to-end translation, has become
competitive with the existing phrase-based statistical machine translation in
many language pairs
\cite{jean2015WMT,Gulcehre-Orhan-et-al-2015,luong2015effective}.


\paragraph{Multilingual Neural Machine Translation}

Existing machine translation systems, mostly based on a phrase-based system or
its variants, work by directly mapping a symbol or a subsequence of symbols in a
source language to its corresponding symbol or subsequence in a target
language. This kind of mapping is strictly specific to a given language {\em
pair}, and it is not trivial to extend this mapping to work on multiple pairs of
languages.

A system based on neural machine translation, on the other hand, can be
decomposed into two modules. The encoder maps a source sentence into a
continuous representation, either a fixed-dimensional vector in the case of the
basic encoder-decoder network or a set of vectors in the case of attention-based
encoder-decoder network. The decoder then generates a target translation based
on this source representation. This makes it possible conceptually to build a
system that maps a source sentence in any language to a common continuous
representation space and decodes the representation into any of the target
languages, allowing us to make a {\em multilingual machine translation} system.

This possibility is straightforward to implement and has been validated in the case of basic
encoder-decoder networks \cite{luong2015multi}. It is however not so, in the case
of the attention-based encoder-decoder network, as the attention mechanism, or
originally called the alignment function in \cite{bahdanau2014neural}, is
conceptually language pair-specific. In \cite{dong2015multi}, the authors
cleverly avoided this issue of language pair-specific attention mechanism by
considering only a one-to-many translation, where each target language decoder
embedded its own attention mechanism. Also, we notice that both of these works
have only evaluated their models on relatively small-scale tasks, making it
difficult to assess whether multilingual neural machine translation can scale
beyond low-resource language translation.

\paragraph{Multi-Way, Multilingual Neural Machine Translation}

In this paper, we first step back from the currently available multilingual
neural translation systems proposed in \cite{luong2015multi,dong2015multi} and
ask the question of whether the attention mechanism can be shared across
multiple language pairs. As an answer to this question, we propose an
attention-based encoder-decoder network that admits a shared attention mechanism
with multiple encoders and decoders. We use this model for all the experiments,
which suggests that it is indeed possible to share an attention
mechanism across multiple language pairs.

The next question we ask is the following: in which scenario would the proposed multi-way,
multilingual neural translation have an advantage over the existing, single-pair
model? Specifically, we consider a case of the translation between a
low-resource language pair. The experiments show that the proposed multi-way,
multilingual model generalizes better than the single-pair translation model,
when the amount of available parallel corpus is small. Furthermore, we validate
that this is not only due to the increased amount of target-side, monolingual
corpus. 

Finally, we train a single model with the proposed architecture on all the
language pairs from the WMT'15; English, French, Czech, German,
Russian and Finnish. 
The experiments show that it is indeed possible to train a single
attention-based network to perform multi-way translation.

\section{Background: Attention-based Neural Machine Translation}
\label{sec2}
The attention-based neural machine translation was proposed in
\cite{bahdanau2014neural}. It was motivated from the observation in
\cite{Cho2014a} that a basic encoder-decoder translation model from
\cite{cho2014learning,sutskever2014sequence} suffers from translating a long
source sentence efficiently. This is largely due to the fact that the encoder of
this basic approach needs to compress a whole source sentence into a single
vector. Here we describe the attention-based
neural machine translation.

Neural machine translation aims at building a single neural network that takes
as input a source sequence $X = \left( x_1, \ldots, x_{T_x} \right)$ and generates a
corresponding translation $Y = \left( y_1, \ldots, y_{T_y} \right)$.  Each symbol in
both source and target sentences, $x_t$ or $y_t$, is an integer index of the
symbol in a vocabulary. 

The encoder of the attention-based model encodes a source sentence into
a set of context vectors $C = \left\{ \vh_1, \vh_2, \ldots, \vh_{T_x} \right\}$,
whose size varies w.r.t. the length of the source sentence.  This
context set is constructed by a bidirectional recurrent neural network (RNN)
which consists of a forward RNN and reverse RNN. The forward RNN reads the
source sentence from the first token until the last one, resulting in the
forward context vectors $\left\{ \ora{\vh}_1, \ldots, \ora{\vh}_{T_x}\right\}$,
where
\vspace{-5px}
\[
    \ora{\vh}_t = \ora{\Psi}_{\text{enc}}\left( \ora{\vh}_{t-1},
    \mE_x\left[x_t\right] \right),
\]
and $\mE_x \in \RR^{|V_x| \times d}$ is an embedding matrix containing row
vectors of the source symbols.  The reverse RNN in an opposite direction,
resulting in $\left\{ \ola{\vh}_1,
\ldots, \ola{\vh}_{T_x}\right\}$, where
\[
    \ola{\vh}_t = \ola{\Psi}_{\text{enc}}\left( \ola{\vh}_{t+1},
    \mE_x\left[x_t\right] \right).
\]

$\ora{\Psi}_{\text{enc}}$ and $\ola{\Psi}_{\text{enc}}$ are recurrent activation
functions such as long short-term memory units (LSTM, \cite{hochreiter1997long})
or gated recurrent units (GRU, \cite{cho2014learning}). 
At each position in the source sentence, the forward and reverse context vectors
are concatenated to form a full context vector, i.e., 
\vspace{-0.25cm}
\begin{align}
    \label{eq:context}
    \vh_t = \left[ \ora{\vh}_t; \ola{\vh}_t \right].
\end{align}

The decoder, which is implemented as an RNN as well, generates one symbol at a
time, the translation of the source sentence, based on the context set returned by
the encoder. At each time step $t$ in the decoder, a time-dependent context
vector $\vc_t$ is computed based on the previous hidden state of the decoder
$\vz_{t-1}$, the previously decoded symbol $\tilde{y}_{t-1}$ and the whole
context set $C$. 

This starts by computing the relevance score of each context vector as
\begin{align}
    \label{eq:score}
    e_{t, i} = f_{\text{score}}(\vh_i, \vz_{t-1}, \mE_y\left[
    \tilde{y}_{t-1}\right]),
\end{align}
for all $i=1,\ldots, T_x$. $\f_{\text{score}}$ can be implemented in various
ways \cite{luong2015effective}, but in this work, we use a simple single-layer
feedforward network. This relevance score measures how relevant the $i$-th
context vector of the source sentence is in deciding the next symbol in the
translation.  These relevance scores are further normalized:
\begin{align}
	\label{eq:normalize}
    \alpha_{t, i} = \frac{\exp(e_{t,i})}{\sum_{j=1}^{T_x} \exp(e_{t, j})},
\end{align}
and we call $\alpha_{t, i}$ the attention weight.

The time-dependent context vector $\vc_t$ is then the weighted sum of the
context vectors with their weights being the attention weights from above:
\begin{align}
	\label{eq:td_context}
    \vc_t = \sum_{i=1}^{T_x} \alpha_{t, i} \vh_i.
\end{align}

With this time-dependent context vector $\vc_t$, the previous hidden state
$\vz_{t-1}$ and the previously decoded symbol $\tilde{y}_{t-1}$, the decoder's
hidden state is updated by
\begin{align}
	\label{eq:decoder}
    \vz_t = \Psi_{\text{dec}}\left( \vz_{t-1}, \mE_y\left[ \tilde{y}_{t-1}
    \right], \vc_t \right),
\end{align}
where $\Psi_{\text{dec}}$ is a recurrent activation function.

The initial hidden state $\vz_0$ of the decoder is initialized based on the
last hidden state of the reverse RNN:
\vspace{-5px}
\begin{align}
    \label{eq:init}
\vz_0 = f_{\text{init}}\left( \ola{\vh}_{T_x} \right),
\end{align}

\noindent where $f_{\text{init}}$ is a feedforward network with one or two hidden layers.

The probability distribution for the next target symbol is computed by
\begin{align}
    \label{eq:target_prob}
    p(y_t = k|\tilde{y}_{< t}, X) \propto
    e^{g_k(\vz_t, \vc_t, \mE\left[ \tilde{y}_{t-1} \right])},
\end{align}
where $g_k$ is a parametric function that returns the unnormalized probability
for the next target symbol being $k$.

Training this attention-based model is done by maximizing the conditional log-likelihood
\begin{align*}
    \LL(\TT) = \frac{1}{N} \sum_{n=1}^N \sum_{t=1}^{T_y} \log p(y_t = y_t^{(n)} |
    y_{<t}^{(n)}, X^{(n)}),
\end{align*}
where the log probability inside the inner summation is from
Eq.~\eqref{eq:target_prob}. It is important to note that the ground-truth target
symbols $y_t^{(n)}$ are used during training. The entire model is differentiable,
and the gradient of the log-likelihood function with respect to all the
parameters $\TT$ can be computed efficiently by backpropagation. This makes it
straightforward to use stochastic gradient descent or its variants to train the
whole model jointly to maximize the translation performance.

\section{Multi-Way, Multilingual Translation}

In this section, we discuss issues and our solutions in extending the
conventional {\em single-pair} attention-based neural machine translation into
{\em multi-way, multilingual} model.

\paragraph{Problem Definition}

We assume $N > 1$ source languages $\left\{ X^1, X^2, \ldots, X^N \right\}$ and
$M > 1$ target languages $\left\{ Y^1, Y^2, \ldots, Y^M\right\}$ , and the
availability of $L \leq M \times N$ {\em bilingual} parallel corpora $\left\{
D_1, \ldots, D_L \right\}$, each of which is a set of sentence pairs of one
source and one target languages. We use $s(D_l)$ and $t(D_l)$ to indicate the
source and target languages of the $l$-th parallel corpus.

For each parallel corpus $l$, we can directly use the log-likelihood function
from Eq.~\eqref{eq:target_prob} to define a pair-specific log-likelihood
$\LL^{s(D_l), t(D_l)}$.  Then, the goal of multi-way, multilingual neural
machine translation is to build a model that maximizes the joint log-likelihood
function
$
{\LL(\TT) = \frac{1}{L} \sum_{l=1}^L \LL^{s(D_l), t(D_l)}(\TT).}
    $
Once the training is over, the model can do translation from any of the source
languages to any of the target languages
included in the parallel corpora.

\subsection{Existing Approaches}


\paragraph{Neural Machine Translation without Attention}

In \cite{luong2015multi}, the authors extended the basic encoder-decoder network
for multitask neural machine translation. As they extended the basic
encoder-decoder network, their model effectively becomes a set of encoders and
decoders, where each of the encoder projects a source sentence into a common
vector space. The point in the common space is then decoded into different
languages.

The major difference between \cite{luong2015multi} and our work is that we
extend the attention-based encoder-decoder instead of the basic model.
This is an important contribution, as the attention-based neural machine
translation has become {\em de facto} standard in neural translation literatures
recently
\cite{jean2014using,jean2015WMT,luong2015effective,sennrich2015neural,sennrich2015improving},
by opposition to the basic encoder-decoder. 

There are two minor differences as well. First, they do not consider
multilinguality in depth. The authors of \cite{luong2015multi} tried only a
single language pair, English and German, in their model. Second, they only
report translation perplexity, which is not a widely used
metric for measuring translation quality. To more easily compare with
other machine translation approaches it would be important to evaluate
metrics such as BLEU, which counts the number of matched
$n$-grams between the generated and reference translations.

\paragraph{One-to-Many Neural Machine Translation}

The authors of \cite{dong2015multi} earlier proposed a multilingual translation
model based on the {\em attention-based neural machine translation}. 
Unlike this paper, they only tried it on one-to-many translation, similarly to
earlier work by \cite{collobert2011natural} where one-to-many natural language
processing was done. In this setting, it is trivial to extend the single-pair
attention-based model into multilingual translation by simply having a single
encoder for a source language and pairs of a decoder and attention mechanism
(Eq.~\eqref{eq:score}) for each target language. We will shortly discuss  more
on why, with the attention mechanism,
one-to-many translation is trivial, while multi-way translation is not.

\subsection{Challenges}

A quick look at neural machine translation seems to suggest a straightforward
path toward incorporating multiple languages in both source and target sides. As
described earlier already in the introduction, the basic idea is simple. We
assign a separate encoder to each source language and a separate decoder to each
target language. The encoder will project a source sentence in its own language
into a common, language-agnostic space, from which the decoder will generate a
translation in its own language. 

Unlike training multiple single-pair neural translation models, in this case,
the encoders and decoders are shared across multiple pairs. This is
computationally beneficial, as the number of parameters grows only linearly with
respect to the number of languages ($O(L)$), in contrary to training separate
single-pair models, in which case the number of parameters grows quadratically
($O(L^2)$.) 

The attention mechanism, which was initially called
a soft-alignment model in \cite{bahdanau2014neural}, aligns a (potentially
non-contiguous) source phrase to a target word. This alignment process is
largely specific to a language pair, and it is not clear whether an alignment
mechanism for one language pair can also work for another pair.

The most naive solution to this issue is to have $O(L^2)$ attention mechanisms
that are {\em not} shared across multiple language pairs. Each attention
mechanism takes care of a single pair of source and target languages. This is
the approach employed in \cite{dong2015multi}, where each decoder had its own
attention mechanism.

There are two issues with this naive approach. First, unlike what has been hoped
initially with multilingual neural machine translation, the number of parameters
again grows quadratically w.r.t. the number of languages. Second and
more importantly, having separate attention mechanisms makes it less
likely for the model to fully benefit from having multiple tasks \cite{caruana1997multitask},
especially for transfer learning towards resource-poor languages.

In short, the major challenge in building a multi-way, multilingual neural
machine translation is in avoiding independent (i.e., quadratically many) attention
mechanisms. There are two questions behind this challenge.
The first one is whether it is even possible to share a single attention
mechanism across multiple language pairs. 
The second question immediately follows: how can we build a
neural translation model to share a single attention mechanism for all the
language pairs in consideration?

\begin{figure}[t]
\centering
\includegraphics[width=.8\columnwidth]{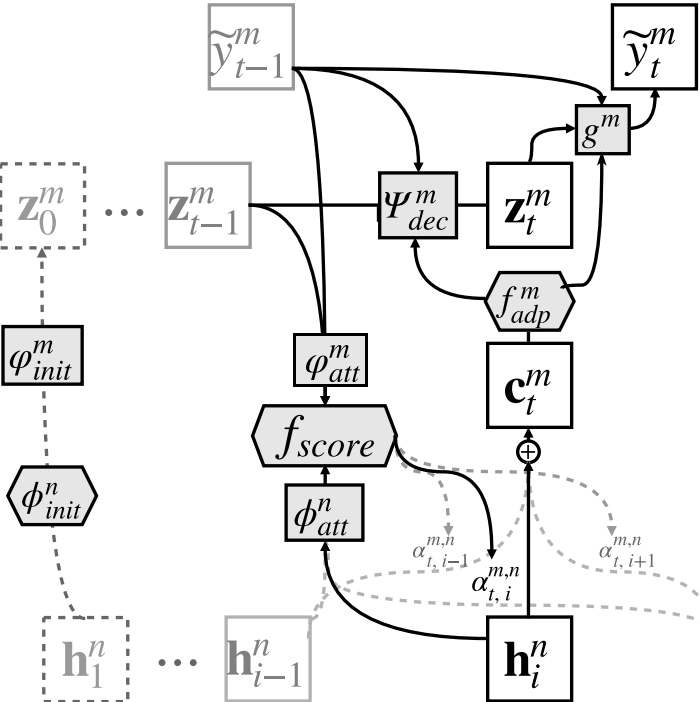}
\caption{One step of the proposed multi-way. multilingual Neural Machine
    Translation model, for the $n$-th encoder and the $m$-th decoder at time
step $t$. See Sec.~\ref{sec:mlnmt} for details.}
\label{fig:shared_att}

    \vspace{-4mm}
\end{figure}

\section{Multi-Way, Multilingual Model}
\label{sec:mlnmt}

We describe in this section a proposed {\em multi-way, multilingual 
attention-based neural machine translation}.  The proposed model consists of $N$
encoders $\{ \Psi_{\text{enc}}^n \}_{n=1}^N$ (see Eq.~\eqref{eq:context}), $M$
decoders $\{ (\Psi_{\text{dec}}^m, g^m, f_{\text{init}}^m) \}_{m=1}^M$ (see
Eqs.~\eqref{eq:decoder}--\eqref{eq:target_prob}) and a shared attention
mechanism $f_{\text{score}}$ (see Eq.~\eqref{eq:score} in the single language pair case).

\paragraph{Encoders}

Similarly to \cite{luong2015effective}, we have one encoder per source language,
meaning that a single encoder is shared for translating the language to multiple
target languages. In order to handle different source languages better, we may
use for each source language a different type of encoder, for instance, of
different size (in terms of the number of recurrent units) or of different
architecture (convolutional instead of recurrent.) This allows
us to efficiently incorporate varying types of languages in the proposed
multilingual translation model.

This however implies that the dimensionality of the context vectors in
Eq.~\eqref{eq:context} may differ across source languages. Therefore, we add to
the original bidirectional encoder from Sec.~\ref{sec2}, a linear transformation
layer consisting of a weight matrix $\mW^n_{\text{adp}}$ and a bias vector
$\vb^n_{\text{adp}}$, which is used to project each context vector into a
common dimensional space:
\vspace*{-3px}
\begin{align}
    \label{eq:ml_context}
    \vh^n_t = \mW^n_{\text{adp}} \left[ \ora{\vh}_t; \ola{\vh}_t \right] +
    \vb_{\text{adp}}^n,
\end{align}
where $\mW^n_{\text{adp}} \in \RR^{d \times (\text{dim}{\ora{\vh}_t} +
\text{dim}{\ola{\vh}_t})}$ and $\vb_{\text{adp}}^n \in \RR^d$.

In addition, each encoder exposes two transformation functions
$\phi_{\text{att}}^n$ and $\phi_{\text{init}}^n$. The first transformer
$\phi_{\text{att}}^n$ transforms a context vector to be compatible with a shared
attention mechanism:
\vspace*{-3px}
\begin{align}
    \label{eq:enc_att}
    \tilde{\vh}_t^n = \phi_{\text{att}}^n(\vh_t^n).
\end{align}
This transformer can be implemented as any type of parametric function, and in
this paper, we simply apply an element-wise $\tanh$ to $\vh_t^n$. 

The second transformer $\phi_{\text{init}}^n$ transforms the first
context vector
$\vh_1^n$ to be compatible with the initializer of the decoder's hidden
state (see Eq.~\eqref{eq:init}): 
\vspace*{-3px}
\begin{align}
    \label{eq:enc_init}
    \hat{\vh}_1^n = \phi_{\text{init}}^n(\vh_1^n).
\end{align}
Similarly to $\phi_{\text{att}}^n$, it can be implemented as any type of
parametric function. In this paper, we use a feedforward network with a single
hidden layer and share one network $\phi_{\text{init}}$ for all encoder-decoder pairs, like attention mechanism.

\begin{table}[t]
    \small
    \centering
    \begin{tabular}{c || c c| c}
        & \multicolumn{2}{c|}{\# Symbols} & \# Sentence \\
        & \# En   & Other& Pairs \\
        \hline
        \hline
En-Fr & 1.022b & 2.213b & 38.85m \\  
En-Cs & 186.57m & 185.58m & 12.12m   \\ 
En-Ru & 50.62m & 55.76m & 2.32m   \\ 
En-De & 111.77m & 117.41m & 4.15m    \\ 
En-Fi & 52.76m & 43.67m & 2.03m  \\ 
    \end{tabular}
    \caption{Statistics of the parallel corpora from WMT'15. Symbols are
        BPE-based sub-words.}
    \label{tab:stats}

    \vspace{-4mm}
\end{table}

\paragraph{Decoders}

We first start with an initialization of the decoder's hidden state. Each
decoder has its own parametric function $\varphi_{\text{init}}^m$ that maps the
last context vector $\hat{\vh}_{T_x}^n$ of the source encoder from
Eq.~\eqref{eq:enc_init} into the initial hidden state:
\begin{align*}
    \vz_0^m = \varphi_{\text{init}}^m(\hat{\vh}_{T_x}^n) =
    \varphi_{\text{init}}^m(\phi_{\text{init}}^n(\vh_{1}^n))
\end{align*}
$\varphi_{\text{init}}^m$ can be any parametric function, and in this paper, we
used a feedforward network with a single $\tanh$ hidden layer.

Each decoder exposes a parametric function $\varphi_{\text{att}}^m$ that
transforms its hidden state and the previously decoded symbol to be compatible
with a shared attention mechanism. This transformer is a parametric function
that takes as input the previous hidden state $\vz_{t-1}^m$ and the previous
symbol $\tilde{y}_{t-1}^m$ and returns a vector for the attention mechanism:
\begin{align}
    \label{eq:dec_att}
    \tilde{\vz}_{t-1}^m = \varphi_{\text{att}}^m\left( \vz_{t-1}^m, \mE_y^m \left[
            \tilde{y}_{t-1}^m
    \right]\right)
\end{align}
which replaces $\vz_{t-1}$ in Eq.~\ref{eq:score}.
In this paper, we use a feedforward network with a single $\tanh$ hidden layer
for each $\varphi_{\text{att}}^m$.

Given the previous hidden state $\vz_{t-1}^m$, previously decoded symbol
$\tilde{y}_{t-1}^m$ and the time-dependent context vector $\vc_t^m$, which we
will discuss shortly, the decoder updates its hidden state:
\begin{align*}
    \vz_t = \Psi_{\text{dec}}\left( \vz_{t-1}^m, \mE_y^m\left[ \tilde{y}^m_{t-1}
    \right], f_{\text{adp}}^m(\vc_t^m) \right),
\end{align*}
where $f_{\text{adp}}^m$ affine-transforms the time-dependent context vector to be of
the same dimensionality as the decoder. We share a single
affine-transformation layer $f_{adp}^m$, for all the decoders in this paper. 

Once the hidden state is updated, the probability distribution over the next
symbol is computed exactly as for the pair-specific model (see
Eq.~\eqref{eq:target_prob}.)

\paragraph{Attention Mechanism}

Unlike the encoders and decoders of which there is an instance for each language,
there is only a single attention mechanism,
shared across all the language pairs. 
This shared mechanism uses the {\em
attention-specific} vectors $\tilde{\vh}_t^n$ and $\tilde{\vz}_{t-1}^m$ from the
encoder and decoder, respectively. 

The relevance score of each context vector $\vh_t^n$ is computed based on the
decoder's previous hidden state $\vz_{t-1}^m$ and previous symbol
$\tilde{y}^m_{t-1}$:
\begin{align*}
    e_{t, i}^{m, n} =& f_{\text{score}}\left(
        \tilde{\vh}_t^n, \tilde{\vz}_{t-1}^m, \tilde{y}^m_{t-1}
    \right) 
\end{align*}
These scores are normalized according to Eq.~\eqref{eq:normalize} to become the
attention weights $\alpha_{t, i}^{m, n}$.

With these attention weights, the time-dependent context vector is computed as
the weighted sum of the {\em original} context vectors:
$
    \vc_t^{m, n} = \sum_{i=1}^{T_x} \alpha_{t, i}^{m, n} \vh_{i}^n.
    $

See Fig.~\ref{fig:shared_att} for the illustration.

\begin{table}[t]
    \small
    \centering
    \begin{tabular}{c c c c c }
        & Size & Single & Single+DF & Multi \\
        \hline
        \hline
        \multirow{4}{*}{\rotatebox[origin=c]{90}{En$\to$Fi}} 
        & 100k & 5.06/3.96 & 4.98/3.99  & 6.2/{\bf 5.17}   \\
        & 200k & 7.1/6.16 & 7.21/6.17  & 8.84/{\bf 7.53}   \\
        & 400k & 9.11/7.85 & 9.31/8.18  & 11.09/{\bf 9.98}   \\
        & 800k & 11.08/9.96 & 11.59/10.15 & 12.73/{\bf 11.28}  \\
        \hline
        \multirow{4}{*}{\rotatebox[origin=c]{90}{De$\to$En}} 
        & 210k  &   14.27/13.2  & 14.65/13.88 & 16.96/{\bf 16.26} \\ 
        & 420k  &   18.32/17.32 & 18.51/17.62 & 19.81/{\bf 19.63} \\ 
        & 840k  &   21/19.93 & 21.69/20.75 & 22.17/{\bf 21.93} \\ 
        & 1.68m & 23.38/23.01 & 23.33/22.86 & 23.86/{\bf 23.52} \\ 
        \hline
        \multirow{4}{*}{\rotatebox[origin=c]{90}{En$\to$De}} 
        & 210k  & 11.44/11.57 & 11.71/11.16 & 12.63/{\bf 12.68} \\
        & 420k  & 14.28/14.25 & 14.88/15.05 & 15.01/{\bf 15.67} \\
        & 840k  & 17.09/17.44 & 17.21/17.88 & 17.33/{\bf 18.14} \\
        & 1.68m & 19.09/19.6  & 19.36/20.13 & 19.23/{\bf 20.59} \\
    \end{tabular}
    \caption{BLEU scores where the target pair's
        parallel corpus is constrained to be 5\%, 10\%, 20\% and 40\% of the
        original size. We report the BLEU scores on the development and test
        sets (separated by /) by the single-pair model (Single), the
        single-pair model with monolingual corpus (Single+DF) and the proposed
    multi-way, multilingual model (Multi).  
}
\label{tab:control}

    \vspace{-4mm}
\end{table}

\section{Experiment Settings}

\subsection{Datasets}

We evaluate the proposed multi-way, multilingual translation model on all the
pairs available from WMT'15--English (En) $\leftrightarrow$ French (Fr), Czech
(Cs), German (De), Russian (Ru) and Finnish (Fi)--, totalling ten directed
pairs. For each pair, we concatenate all the available parallel corpora from
WMT'15 and use it as a training set. We use newstest-2013 as a development set
and newstest-2015 as a test set, in all the pairs other than
Fi-En. In the case of Fi-En, we use newsdev-2015 and newstest-2015 as a
development set and test set, respectively. 

\paragraph{Data Preprocessing}

Each training corpus is tokenized using the tokenizer script from the Moses decoder.
The tokenized training corpus is cleaned following the procedure in
\cite{jean2015WMT}. Instead of using space-separated tokens, or words, we use
sub-word units extracted by byte pair encoding, as recently proposed in
\cite{sennrich2015neural}. For each and every language, we include 30k sub-word
symbols in a vocabulary. See Table~\ref{tab:stats} for the statistics of the
final, preprocessed training corpora.

\paragraph{Evaluation Metric}

We mainly use BLEU as an evaluation metric using the multi-bleu script from
Moses.
BLEU is computed on the tokenized text after merging the BPE-based sub-word
symbols. We further look at the average log-probability assigned to reference
translations by the trained model as an additional evaluation metric, as a way
to measure the model's density estimation performance free from any error caused
by approximate decoding.



\subsection{Two Scenarios}

\paragraph{Low-Resource Translation}
First, we investigate the effect of
the proposed multi-way, multilingual model on low-resource language-pair
translation. Among the five languages from WMT'15, we choose En, De and Fi as
source languages, and En and De as target languages. We control the amount of
the parallel corpus of each pair out of three to be 5\%, 10\%, 20\% and 40\% of
the original corpus. In other words, we train four models with different sizes
of parallel corpus for each language pair (En-De, De-En, Fi-En.) 

As a baseline, we train a single-pair model for each multi-way, multilingual
model. We further finetune the single-pair model to incorporate the target-side
monolingual corpus consisting of all the target side text from the other
language pairs (e.g., when a single-pair model was trained on Fi-En, the
target-side monolingual corpus consists of the target sides from De-En.) This is
done by the recently proposed deep fusion \cite{Gulcehre-Orhan-et-al-2015}. The
latter is included to tell whether any improvement from the multilingual model
is simply due to the increased amount of target-side monolingual corpus.

\paragraph{Large-scale Translation}
We train one multi-way, multilingual model that has five
encoders and five decoders, corresponding to the five languages from WMT'15; En,
Fr, De, Cs, Ru, Fi $\to$ En, Fr, De, Cs, Ru, Fi. We use the full corpora for all
of them.

\subsection{Model  Architecture} 
Each symbol, either source or target, is projected on a 620-dimensional space.
The encoder is a bidirectional recurrent neural network with 1,000 gated
recurrent units (GRU) in each direction, and the decoder is a recurrent neural network with
also 1,000 GRU's. The decoder's output function $g_k$ from
Eq.~\eqref{eq:target_prob} is a feedforward network with 1,000 $\tanh$ hidden
units. The dimensionalities of the context vector $\vh_t^n$ in
Eq.~\eqref{eq:ml_context}, the attention-specific context vector
$\tilde{\vh}_t^n$ in Eq.~\eqref{eq:enc_att} and the attention-specific decoder
hidden state $\tilde{\vh}_{t-1}^m$ in Eq.~\eqref{eq:dec_att} are all set to
1,200.

We use the same type of encoder for every source language, and the same type of
decoder for every target language.  The only difference between the single-pair
models and the proposed multilingual ones is the numbers of encoders $N$ and
decoders $M$. We leave those multilingual translation specific components, such
as the ones in Eqs.~\eqref{eq:ml_context}--\eqref{eq:dec_att}, in the
single-pair models in order to keep the number of shared parameters constant.

%

\begin{table*}[ht]
    \small
    \centering
    \begin{minipage}{\textwidth}
        \centering
        \begin{tabular}{c | c c || c c |  c c |  c  c |  c c |  c c}
            \multicolumn{3}{r||}{} 
            & \multicolumn{2}{c|}{Fr (39m)} 
            & \multicolumn{2}{c|}{Cs (12m)} 
            & \multicolumn{2}{c|}{De (4.2m)} 
            & \multicolumn{2}{c|}{Ru (2.3m)} 
            & \multicolumn{2}{c}{Fi (2m)}  \\
            \cline{4-13}
            \multicolumn{3}{r||}{Dir} 
            & $\to$ En & En $\to$ 
            & $\to$ En & En $\to$ 
            & $\to$ En & En $\to$ 
            & $\to$ En & En $\to$ 
            & $\to$ En & En $\to$  \\
            \hline
            \hline
            \multirow{4}{*}{\rotatebox[origin=c]{90}{(a) BLEU}}
            &
            \multirow{2}{*}{\rotatebox[origin=c]{90}{Dev}}
            & Single & 27.22 & 26.91 & 21.24 & 15.9 & 24.13 & 20.49 & 21.04 & 18.06 & 13.15 & 9.59 \\
            &
            & Multi & 26.09 & 25.04 & 21.23 & 14.42 & 23.66 & 19.17 & 21.48 & 17.89 & 12.97 & 8.92 \\
            \cline{2-13}
            &
            \multirow{2}{*}{\rotatebox[origin=c]{90}{Test}}
            & Single & 27.94 & {\bf 29.7} & 20.32 & {\bf 13.84} & 24 &
            {\bf 21.75} & 22.44 & {\bf 19.54} & 12.24 & {\bf 9.23} \\
            &
            & Multi & {\bf 28.06} & 27.88 & {\bf 20.57} & 13.29 & {\bf 24.20} & 20.59 & {\bf 23.44}& 19.39 & {\bf 12.61} & 8.98 \\
            \hline
            \hline
            \multirow{4}{*}{\rotatebox[origin=c]{90}{(b) LL}}
            &
            \multirow{2}{*}{\rotatebox[origin=c]{90}{Dev}}
            & Single                & -50.53          & -53.38         & -60.69                               & -69.56         & -54.76          & -61.21 & -60.19          & -65.81         & -88.44          & -91.75         \\
            &
            & Multi                 & -50.6          & -56.55         & -54.46                               & -70.76         & -54.14          & -62.34 & -54.09          & -63.75         & -74.84          & -88.02         \\ 
            \cline{2-13}
            &
            \multirow{2}{*}{\rotatebox[origin=c]{90}{Test}}
            & Single & -43.34 & {\bf -45.07} & -60.03 & {\bf -64.34} & -57.81 & {\bf -59.55} & -60.65 & -60.29 & -88.66 & -94.23 \\
            & 
            & Multi  & {\bf -42.22} & -46.29 & {\bf -54.66} & -64.80 & {\bf -53.85} & -60.23 & {\bf -54.49} & {\bf -58.63} & {\bf -71.26} & {\bf -88.09} \\
        \end{tabular}
    \end{minipage}

    \caption{(a) BLEU scores and (b) average log-probabilities for all the five
        languages from WMT'15. 
    }
    \label{tab:all}

    \vspace{-4mm}
\end{table*}

\subsection{Training}
\label{sec:training}

\paragraph{Basic Settings}
We train each model using stochastic gradient descent (SGD) with Adam
\cite{kingma2014adam} as an adaptive learning rate algorithm. We use the initial
learning rate of $2\cdot 10^{-4}$ and leave all the other hyperparameters as
suggested in \cite{kingma2014adam}. Each SGD update is computed using a
minibatch of 80 examples, unless the model is parallelized over two GPUs, in
which case we use a minibatch of 60 examples. We only use sentences of length up
to 50 symbols. We clip the norm of the gradient to be no more than $1$
\cite{pascanu2012difficulty}.  All training runs are early-stopped based on BLEU
on the development set. As we observed in preliminary experiments better scores on the 
development set when finetuning the shared
parameters and output layers of the decoders in the case of multilingual models,
we do this for all the multilingual models. During
finetuning, we clip the norm of the gradient to be no more than $5$.

%

\paragraph{Schedule} 

As we have access only to bilingual corpora, each sentence pair updates only a
subset of the parameters. Excessive updates based on a single language pair may
bias the model away from the other pairs. To avoid it, we cycle through all the
language pairs, one pair at a time, in Fi$\leftrightarrows$En, De$\leftrightarrows$En, Fr$\leftrightarrows$En, Cs$\leftrightarrows$En, Ru$\leftrightarrows$En order.\footnote{$\leftrightarrows$ indicates simultaneous updates on two GPUs.} 


\paragraph{Model Parallelism}

The size of the multilingual model grows linearly w.r.t. the number of
languages. We observed that a single model that handles five source and five
target languages does not fit in a single GPU 
during training. We address this by 
distributing computational paths according to different translation pairs over
multiple GPUs.
The shared parameters, mainly related to the attention mechanism, is duplicated
on both GPUs. 
The implementation was based on the work in \cite{DingWMT14}.


\section{Results and Analysis}

\paragraph{Low-Resource Translation}

It is clear from Table~\ref{tab:control} that the proposed model (Multi)
outperforms the single-pair one (Single) in all the cases. This is true even
when the single-pair model is strengthened with a target-side monolingual corpus
(Single+DF). This suggests that the benefit of generalization from having
multiple languages goes beyond that of simply having more target-side
monolingual corpus. The performance gap grows as the size of target parallel
corpus decreases.

%
%

\paragraph{Large-Scale Translation}

In Table~\ref{tab:all}, we observe that the proposed multilingual model
outperforms or is comparable to the single-pair models for the majority of the
all ten pairs/directions considered. This happens in terms of both BLEU and
average log-probability. This is encouraging, considering that there are twice
more parameters in the whole set of single-pair models than in the
multilingual model.

It is worthwhile to notice that the benefit is more apparent when the model
translates from a foreign language to English. This may be due to the fact that
all of the parallel corpora include English as either a source or target
language, leading to a better parameter estimation of the English decoder. In
the future, a strategy of using a pseudo-parallel corpus to increase the amount
of training examples for the decoders of other languages
\cite{sennrich2015improving} should be investigated to confirm this conjecture.




\section{Conclusion}
\vspace{-5px}
In this paper, we proposed multi-way, multilingual attention-based neural
machine translation. The proposed approach allows us to build a single neural
network that can handle multiple source and target languages simultaneously.
The proposed model is a step forward from the recent works on multilingual
neural translation, in the sense that we support attention mechanism, compared
to \cite{luong2015multi} and multi-way translation, compared to
\cite{dong2015multi}. Furthermore, we evaluate the proposed model on large-scale
experiments, using the full set of parallel corpora from WMT'15.

We empirically evaluate the proposed model in large-scale experiments using all
five languages from WMT'15 with the full set of parallel corpora and also in the
settings with artificially controlled amount of the target parallel corpus. In
both of the settings, we observed the benefits of the proposed multilingual
neural translation model over having a set of single-pair models. The
improvement was especially clear in the cases of translating low-resource
language pairs.

We observed the larger improvements when translating to
English. We conjecture that this is due to a higher availability of English in
most parallel corpora, leading to a better parameter estimation of the English
decoder. More research on this phenomenon in the future will result in further
improvements from using the proposed model.  Also, all the other techniques
proposed recently, such as ensembling and large vocabulary tricks, need to be
tried together with the proposed multilingual model to improve the translation
quality even further. Finally, an interesting future work is to use the proposed
model to translate between a language pair not included in a set of training
corpus.

\section*{Acknowledgments}

The authors would like to thank the developers of Theano \cite{bergstra2010,Bastien2012} and Blocks \cite{MerrienboerBDSW15}. We acknowledge the support of the following organizations for research funding and computing support: NSERC, Samsung, IBM, Calcul Qu\'ebec, Compute Canada, the Canada Research Chairs, CIFAR and TUBITAK-2214a.

%

\bibliography{mlnmt}
\bibliographystyle{naaclhlt2016}

\end{document}